\documentclass{article}
\usepackage{spconf,amsmath,graphicx}
\usepackage[english]{babel}
\usepackage[utf8x]{inputenc}
\usepackage{amsmath}
\usepackage{amssymb}
\usepackage{graphicx}
\usepackage[colorinlistoftodos]{todonotes}
\usepackage[margin=3cm]{geometry}
\usepackage{hyperref}
\usepackage[numbers]{natbib}
\usepackage{fancyhdr}
\usepackage{enumitem}
\usepackage{subcaption}
\usepackage[toc,page]{appendix}
\usepackage{caption}
\usepackage{pbox}
\usepackage{listings}             
\usepackage{hyperref}
\usepackage{datetime}
\usepackage{color}

\usepackage{color}

\DeclareMathOperator*{\argmin}{arg\,min}

\title{Graph Learning under Sparsity Priors}
\name{Hermina Petric Maretic, Dorina Thanou and Pascal Frossard}
\address{Signal Processing Laboratory (LTS4),  EPFL, Switzerland }
\begin{document}
\ninept
\maketitle
\begin{abstract}
Graph signals offer a very generic and natural representation for data that lives on networks or irregular structures. The actual data structure is however often unknown a priori but can sometimes be estimated from the knowledge of the application domain. If this is not possible, the data structure has to be inferred from the mere signal observations. This is exactly the problem that we address in this paper, under the assumption that the graph signals can be represented as a sparse linear combination of a few atoms of a structured graph dictionary. The dictionary is constructed on polynomials of the graph Laplacian, which can sparsely represent a general class of graph signals composed of localized patterns on the graph. We formulate a graph learning problem, whose solution provides an ideal fit between the signal observations and the sparse graph signal model. As the problem is non-convex, we propose to solve it by alternating between a signal sparse coding and a graph update step. We provide experimental results that outline the good graph recovery performance of our method, which generally compares favourably to other recent network inference algorithms. % \pascal{revise when we have all results...}
\end{abstract}
\section{Introduction}
\label{sec:intro}
Graphs provide a flexible tool for modelling and manipulating complex data that resides on topologically complicated domains such as transportation networks, social, and  computer networks, brain analysis or even digital images. Typically, once a graph is well-defined,  classical data analysis and inference tasks can be effectively performed by using tools such as spectral graph theory or generalization of signal processing notions in the graph domain \cite{Shuman13a}.    %In many of the above mentioned examples, graph structures emerge naturally. 
However, it is often the case that the graph structures are not defined a priori or the intrinsic relationships between different signal observations are not clear or are defined only in a very subtle manner. Since the definition of a meaningful graph plays a crucial role in the analysis and processing of structured data, the inference of the graph itself from the data is a very important problem that has not been well investigated so far.

%In many examples in nature and society, such as geographical, traffic or social networks, graph structures are naturally imposed. However, not only are these graph structures not uniquely defined (we might want to observe connections on a social network as friendship connections, but also as connections of people that share similar interests or connections of people that live close to each other etc.); but in some cases the intrinsic relationship between different observations is not clear or is defined only in a very subtle manner.

%At the same time, the emerging field of signal processing on graphs studies tools for analysis of signals defined on graph vertices, modelling the pairwise relationship between them through edge weights \cite{Shuman13a}. It is then clear that observing our signals on a meaningful graph plays a crucial role in this analysis. For this reason we explore the construction of relevant graphs and the possibilities of learning them from signals.

The definition of the graph has been initially studied in a machine learning framework with 
%with the aim of using the extra information provided by the graph in analyzing data similarities and solving standard machine learning problems. These type of works usually rely on 
simple models such as K-nearest neighbour graphs and other heuristically defined graph kernels \cite{Ng02}, \cite{Zhou04} or convex combinations of them \cite{Argyriou}. 
%Creating convex combinations of these simple graphs while giving preference to better ones, has also been an approach in constructing a meaningful topology \cite{argyriou}. 
Richer adaptivity to the data is obtained by relating the graph structure more closely to data properties in order to infer a graph topology. Techniques such as sparse inverse covariance estimations \cite{Banerjee08}, \cite{Friedman08} rely on Gaussian graphical models to identify partial correlations between random variables and define graph edges. Such techniques have also been extended to very large graphs, by learning highly structured representations %\cite{toh},
 \cite{celik14}. More recent works relax the assumption of a full rank precision matrix by infering the graph topology from a graph signal processing perspective \cite{Dong_2015}, \cite{Kalofolias16}, \cite{Pavez16}, \cite{PasdeloupGMPR16} under explicit graph signal smoothness assumptions. The above works assume that the data globally evolves smoothly on the underlying structure. However, such a model might not be very precise for many real world  datasets, which can feature highly localized behaviors or piecewise smoothness. The recent framework in \cite{SegarraMMR16} that observes graph signals as white signals filtered with a graph shift operator polynomial, is one of the first network inference works to depart from explicit global smoothness assumptions. A similar idea uses adjacency matrix polynomials to model causation in time-varying signals \cite{mei2015signal}, \cite{mei2016signal}.

In this work, we consider a generic model where the graph signals are represented by (sparse) combinations of overlapping local patterns that reside on the graph. That is, given a set of graph signals, we model these signals as a linear combination of only a few components (i.e., atoms) from a graph structured dictionary that captures localized patterns on the graph. We incorporate the underlying graph structure into the dictionary through the graph Laplacian operator. In order to ensure that the atoms are localized in the graph vertex domain, we further impose the constraint that our dictionary is a concatenation of subdictionaries that are polynomials of the graph Laplacian \cite{Thanou14}. Based on this generic model, we cast a new graph learning problem that aims at estimating a graph that explains the data observations, which should eventually form a sparse set of localized patterns on the learned graph.  We propose an alternating optimization algorithm to address the resulting nonconvex inverse problem, which is based on a sparse coding step obtained with orthogonal matching pursuit (OMP) and a graph learning step performed by applying a projected gradient descent step. We finally provide a few illustrative experiments on synthetic data, and we show that our generic graph signal model leads to better graph recovery performance than state-of-the-art algorithms. %\pascal{revise based on final set of results}

The rest of the paper is organized as follows. We describe our graph learning framework in Section 2. We then present our alternating optimization algorithm in Section 3, and evaluate the graph recovery performance in Section 4.

%Finally, we We impose the atoms to be polynomials of the graph Laplacian 
%we infer a meaningful graph structure by exploiting the  sparse representation of these signals in a structured graph dictionary that captures the localized patterns. 

%and a sparse representation of signals on that graph. In particular, we assume signals are generated from a small number of atoms in an overcomplete dictionary and model this as an optimisation problem. We propose a solution through orthogonal maching pursuit (OMP) and gradient descent and show our algorithm performs better than the state-of-the-art on such problems.

%\subsection{Related work}

\section{Graph Learning Framework}

\subsection{Sparse signal representation on graphs}

We first present the graph signal processing framework used in this work. We consider an undirected, weighted graph $G = (V, E, W)$ with a set of $N$ vertices $V$, edges $E$ and a weighted adjacency matrix $W$, with the weight value $W_{ij}=0$ if there is no edge between $i$ and $j$. We define a signal on the graph $G$ as a function $y:V \rightarrow \mathbb{R}$, where $y(v)$ denotes the value of a signal on a vertex $v$. %We consider this signal smooth if the values associated with vertices connected with high weighted edges tend to be similar. 
One of the most useful graph operators that is widely used in graph signal processing tasks, is %non-normalized (or combinatorial) graph Laplacian, defined as 
%\begin{align*}
%L = D - W,
%\end{align*} 
%where $D$ is the diagonal degree matrix and i
the normalized Laplacian operator, defined as 
\begin{align*}
\mathcal{L} = I - D^{-\frac{1}{2}}WD^{-\frac{1}{2}},
\end{align*} where $I$ is the identity matrix and $D$ is the diagonal degree matrix. This graph operator has a complete set of orthonormal eigenvectors $\chi = \{\chi_0,\chi_1, ..., \chi_{N-1}\}$  with a corresponding set of non-negative eigenvalues. These eigenvectors form a basis for the definition of the graph Fourier transform \cite{Shuman13a}, which provides a spectral representation of graph signals.

Similarly to classical signal processing, one can design an overcomplete dictionary $\mathcal{D}$ for signals on graphs, such that  every graph signal $y$ %signals on the graph can be sparsely represented through dictionary atoms, 
can be represented as a sparse linear combination of dictionary atoms, i.e., $y \approx \mathcal{D} x $, 
%\begin{align*}
%y = \mathcal{D}x, 
%\end{align*}
where $x$ is a sparse vector of coefficients. In order to obtain an effective representation of graph signals, the dictionary has to incorporate the structure of the graph. %the dictionary has to carry the the graph. %A parametric dictionary is a good choice for this problem, as it captures both the structure of the graph, but also the training signals at hand, and has a computationally inexpensive implementation \cite{dictionaries}. 
In particular, we consider here a polynomial graph structured  dictionary, which has been shown to provide sparse representations of a general class of graph signals that are linear combinations of graph patterns positioned at different vertices of the graph \cite{Thanou14}. The dictionary 
$\mathcal{D} = \left[\mathcal{D}_1, \mathcal{D}_2, ..., \mathcal{D}_S\right]$ is defined as a concatenation of $S$ subdictionaries of the form
\begin{align*}
\mathcal{D}_s &= \hat{g}_s (\mathcal{L}) = \sum_{k=0}^K \alpha_{sk} \mathcal{L}^k \\
&= \sum_{k=0}^K \alpha_{sk} (I - D^{-\frac{1}{2}}WD^{-\frac{1}{2}})^k,
\end{align*}
where $\hat{g}_s(\cdot)$ is the generating kernel of the subdictionary $\mathcal{D}_s$. In this particular signal model, the graph Laplacian captures the connectivity of the graph, while the polynomial coefficients reflect the distribution of the signal in particular neighbourhoods of the graph. Furthermore, the behaviour of these kernels in graph eigenvalues describes the nature of atom signals in the graph. Namely, a kernel promoting low frequency components will result in smooth atoms on the graph. The polynomial coefficients, together with the graph Laplacian, fully characterize the dictionary. In the rest of this paper, we consider the general class of graph signals that have sparse representation in the dictionary $\mathcal{D}$. 

%Furthermore, zero appears as an eigenvalue with multiplicity equal to the number of connected components of the graph and the spectrum of the normalised graph Laplacian satisfies
%\begin{align*}
%\sigma (\mathcal{L}) = \{0 = \lambda_0 \leq \lambda_1 \leq ... \leq \lambda_{N-1} \leq 2\}
%\end{align*}
%This leads to observation of graph signals in the graph spectral domain (as opposed to the vertex domain). Similarly to the classical Fourier transform, we can define the graph Fourier transform $\hat{y}$ of a signal $y$ at frequency $\lambda_l$ as the expansion:
%\begin{align*}
%\hat{y} (\lambda_l) = \left\langle y, \chi_l \right\rangle = \sum_{n=1}^N y(n) \chi_l^{*}(n),
%\end{align*}
%and the inverse graph Fourier transform as
%\begin{align*}
%y(n) = \sum_{l=0}^{N-1} \hat{y} (\lambda_l) \chi_l (n).
%\end{align*}
%Here, the normalised Laplacian eigenvectors form a Fourier basis and it is not difficult to see that the corresponding eigenvalues carry a notion of frequency.

\subsection{Problem formulation}

Equipped with the sparse graph signal model defined above, we can now formulate our graph learning problem. In particular, 
given a set of signal observations $Y = \left[y_1, y_2, ..., y_M\right] \in \mathbb{R}^{N\times M}$, we want to infer a graph $G$, such that the observed signals have a sparse representation in the graph dictionary built on $G$. 
%We would therefore like to minimise the difference between the signals and their representation in the given dictionary $||Y - \mathcal{D}X||_F^2$, assuming the sparsity of the representation, but also the sparsity of the graph.

More formally, the graph learning problem can be cast as follows:
\begin{align}
 \nonumber \argmin_{W, X} ~~~ &||Y - \mathcal{D}X||_F^2 + \beta_W ||W||_1\\
\text{subject to }
\nonumber~~&\mathcal{D} = \left[\mathcal{D}_1, \mathcal{D}_2, ..., \mathcal{D}_S\right],\\
\nonumber~~~ &\mathcal{D}_s = \sum_{k=0}^K \alpha_{sk} \mathcal{L}^k, \forall s\in \{1, ..., S\}\\
\label{opt_prob}~~~ &\mathcal{L} = I - D^{-\frac{1}{2}}WD^{-\frac{1}{2}}\\
\nonumber~~~ &W_{ij} = W_{ji} \geq 0, \forall
i,j, i\neq j\\
\nonumber~~~ &W_{ii} = 0, \forall i \\
\nonumber~~~ &||x_m||_0 \leq T_0, \forall m\in \{1, ..., M\}
\end{align}
%\pascal{format the above problem properly}
where $T_0$ is the sparsity level of the coefficients of each signal, $\beta_W$ is a parameter that controls the graph sparsity i.e., the number of non-zero edges of the graph, through the $L_1$ norm of the graph adjacency matrix $W$, and $x_m$ is the $m^{th}$ column of the matrix $X$. The optimization is performed over the weight matrix $W$ instead of $\mathcal{L}$ that is used explicitly in the dictionary construction, as the constraints defining a valid weight matrix $W$ are much simpler to handle than those defining a valid Laplacian. Namely, a valid $\mathcal{L}$ must be positive semi-definite, while the weight matrix assumes only symmetry and non-negativity of the weights. 

Finally, we assume in this work that the dictionary kernels, i.e., the coefficients $\alpha_{sk}$, are known. We can for example model these generating kernels $\hat{g}_s(\cdot)$ as graph heat kernels and compute the polynomial coefficients $\alpha_{sk}$ as a $K$-order Taylor approximation coefficients. Other kernel choices are possible, such as spectral graph wavelets \cite{Hammond11}, where the polynomial coefficients could be inferred from a Chebyshev polynomial approximation, or a priori learned kernels \cite{Thanou14}. For the sake of simplicity, we also consider the $S$ and $K$ are determined by a priori information about the nature of the target application, or optimized separately. 

\section{Graph learning algorithm}

Next, we discuss the solution obtained by our graph learning framework. As the optimization problem  (\ref{opt_prob}) is non-convex, we solve the problem by alternating between the sparse coding and the weight matrix update steps, which is a widely used techniques for solving ill-posed non-convex problems. 

%\subsection{Estimation of $X$}
In the first step, we fix the weight matrix $W$ and optimize the objective function with respect to the sparse codes $X$, which leads to the following optimization problem 
\begin{align*}
\argmin_{X} &||Y - \mathcal{D}X||_F^2\\
\text{subject to } &||x_m||_0 \leq T_0, \forall m\in \{1, ..., M\}.
\end{align*}
The  $L_0$-``norm" constraint ensures sparsity of the sparse codes. 
We solve the above problem using orthogonal matching pursuit (OMP) \cite{Tropp04}. Before updating $X$, we normalize the atoms of the dictionary $\mathcal{D}$ to be of unit norm.  This step is essential for the OMP step in order to treat all atoms equally. To recover our initial structure, after computing $X$, we renormalize the atoms of our dictionary and the sparse coding coefficients in such a way that the product $\mathcal{D} X$ remains constant. % Other methods yielding less accurate, but faster results, can be used. This would include matching pursuit (MP) or incorporating a convex $L_1$ relaxation of  the $L_0$ constraint.
%\subsection{Estimation of $W$}

In the second step, we fix the sparse codes and we update the dictionary, i.e., the graph. Estimating the weight matrix with fixed sparse codes $X$, however, remains non-convex, as the dictionary $\mathcal{D}$ is constructed from a $K$-order polynomials of $\mathcal{L}$, and thus of $W$. The optimization problem becomes:
\begin{align*}
\argmin_{W}& ||Y - \mathcal{D}X||_F^2 + \beta_W ||W||_1\\
\text{subject to }
\nonumber~~&\mathcal{D} = \left[\mathcal{D}_1, \mathcal{D}_2, ..., \mathcal{D}_S\right],\\
\nonumber~~~ &\mathcal{D}_s = \sum_{k=0}^K \alpha_{sk} \mathcal{L}^k, \forall s\in \{1, ..., S\}\\
~~~ &\mathcal{L} = I - D^{-\frac{1}{2}}WD^{-\frac{1}{2}}\\
~~~ & W_{ij} = W_{ji} \geq 0, \forall
i,j, i\neq j\\
~~~ & W_{ii} = 0, \forall i.
\end{align*}
We propose a solution based on a gradient descent step followed by simple projections into the weight matrix constraints.  The gradient of the smooth term of the objective function can be given in a closed form as follows
\begin{align*}
\nonumber &\nabla_W\|Y -\sum_{s=1}^S\mathcal{D}_s X_s\|_{F}^2 \\ 
& = \sum_{s=1}^S \sum_{k=1}^K\alpha_{sk}\big(-\sum_{r = 0}^{k-1} 2 A_{k,r}^T + \mathbf{1}_{N\times N}(B_{k}\circ I) \big), 
\end{align*}
where 
$A_{k,r} = D^{-1/2}\mathcal{L}^{k-r-1} X_s (Y - \mathcal{D}X)^T \mathcal{L}^r D^{-1/2},$
$B_{k} = \sum_{r=0}^{k-1}D^{-1/2}WA_{k,r}D^{-1/2} + A_{k,r} W D^{-1},$ $\mathbf{1}_{N\times N}$ is a matrix of ones, $I$ is an $N\times N$ identity matrix, and $\circ$ denotes the pairwise (Hadamard)  product. This result is obtained by using properties of the trace operator and applying the chain rule for the gradient. However, we omit the detailed derivation of the gradient due to space constraints.  
%Note that the gradient of $ ||Y - \mathcal{D}X||_F^2$ is given with:
%\begin{align*}
%\frac{\partial}{\partial W} \|Y-\mathcal{D}X\|_F^2 = \\
%\frac{\partial}{\partial W} tr((Y-\sum_s \sum_k \alpha_{sk} \mathcal{L}^k X_s)(Y-\sum_s \sum_k \alpha_{sk} \mathcal{L}^k X_s)^T)=\\
%\end{align*}
%while we use a proximal operator to approximate the gradient of the non-differentiable $\beta_W ||W||_1$.
To approximate the gradient of the non-differentiable $\beta_W ||W||_1$ term, we use $\beta_W \mbox{sign}(W)$. This operation ends up shrinking all positive elements by $\beta_W$ in every step, while not changing the zero ones. 
To avoid complex projections, we use a symmetric gradient with a zero-diagonal. By doing so, we are optimizing over only $N(N-1)/2$ variables, instead of $N^2$. Note that the projections are quite simple, i.e.,  
\begin{align*}
\underset{\tilde{W_{ij}}\geq 0}{\text{argmin} }||\tilde{W_{ij}} - W_{ij}||_F^2 = 
\begin{cases}
    0,& \text{if } W_{ij}\leq 0\\
    W_{ij},              & \text{otherwise,}
\end{cases}
\end{align*}
and even promote sparsity. The way that we approximate the gradient of the $\beta_W ||W||_1$ term,   together with the projection to the positive space, bears strong  similarity to using  the proximal  operator for the $L_1$ norm, with the only  difference being in the fact that we project to the positive space after performing  the  gradient descent.  It is important to note that since we are alternating between the sparse coding and the graph update step, there is no theoretical guarantee for convergence to a local optimum. However,  the method has shown promising results in all conducted tests, as we will see in the next section. %However, it is important to note that this method has no theoretical convergence guaranties.

\begin{figure}[t]
\centering
\includegraphics[scale=0.3]{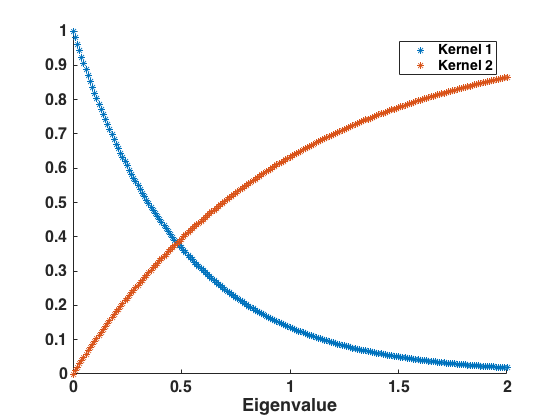}
\caption{Generating kernels for a polynomial dictionary}
\label{fig:kernels}
\end{figure}

\section{Simulation results}
%In this section, we quantify the performance of our algorithm in both synthetic and real world data. 
%\subsection{Synthetic results}

\begin{figure}[b!]
        \begin{subfigure}[b]{0.22\textwidth}   
            \centering 
            \includegraphics[width=\textwidth]{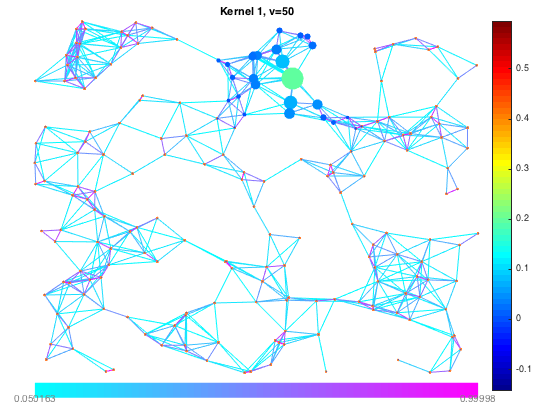}
            \caption{Kernel 1}
        \end{subfigure}
        \quad
        \begin{subfigure}[b]{0.22\textwidth}   
            \centering 
            \includegraphics[width=\textwidth]{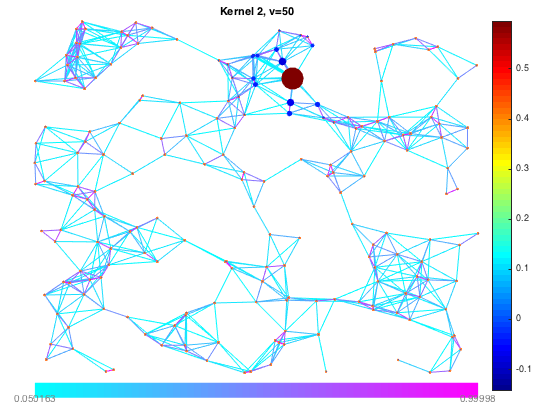}
            \caption{Kernel 2}
        \end{subfigure}
        \caption
        {\small An example of atoms generated from two different polynomial kernels and centered at the same vertex.} 
        \label{fig:atoms}
\end{figure}

We have tested the performance of our algorithms on synthetic data that follow our signal model. We carry our experiments on sparse Erd\H{o}s-R\'{e}nyi model (ER) \cite{Erdos60} graphs with $N=20, 50$ and 100 vertices, and on an radial basis function (RBF) random graph. In the case of the RBF graph, we generate the coordinates of the vertices uniformly at random in the unit square, and we set the edge weights based on a thresholded Gaussian kernel function:
\begin{align*}
W(i,j)=\begin{cases}
      e^{-\frac{[\text{dist}(i,j)]^2}{2\sigma^2}}, & \text{if}\ dist(i,j)\leq \kappa \\
      0, & \text{otherwise}
    \end{cases}
\end{align*}
All graph topologies are designed to have approximately $3N$ edges, and we generate 100 random instances of the graph. For numerical reasons, every graph is stripped of its isolated vertices, but full graph connectedness is not necessarily insured. For each instance of the graph, we then construct a parametric dictionary as a concatenation of $S = 2$ subdictionaries designed as polynomials of degree $K = 15$. The generating kernels for each subdictionary are defined by the Taylor approximation of two heat kernels, one of which is shifted in the spectral domain to cover high graph frequencies, allowing for a more general class of signals. More precisely, we define 
\begin{align*}
\hat{g}_1(\lambda) \approx e^{-2 \lambda}, \quad \hat{g}_2(\lambda) \approx 1 - e^{- \lambda},
\end{align*} 
where we use $\approx$ because of the fact that each exponential function is approximated by a 15-order polynomial. 
The obtained kernels are illustrated in Fig. \ref{fig:kernels}. An example of corresponding atoms for a graph with a thresholded Gaussian kernel can be seen on Fig. \ref{fig:atoms}.

Then, we generate a set of 200 training graph signals using randomly generated sparse coding coefficients from a normal distribution. These sparse codes represent each signal as a linear combination of $T_0 = 4$ atoms from the dictionary. We use these training signals and the known polynomial coefficients to learn a graph with our proposed graph learning algorithm. The weight matrix is initialized as a random symmetric matrix with values between 0 and 1, and a zero diagonal. The parameter $\beta_W$ and the gradient descent step size are determined with grid search.

We threshold the small entries of the learned matrix  in order to recover a strongly sparse graph structure. Determining a good threshold value proves to be crucial in obtaining a relevant graph. As a rule of thumb in these synthetic settings, we threshold our results in such a way that the number of learned edges approximately equals to the number of the edges of the groundtruth graph. In more general settings, where the groundtruth graph is not known, we discard the entries of the matrix that are smaller than a predefined threshold of $10^{-4}$. 

\begin{table}[h!]
\centering
\caption{Mean accuracies for different graph sizes}
\begin{tabular}{|c|c|c|c|}
\hline 
mean value/graph size & \textbf{20} & \textbf{50} & \textbf{100} \\ 
\hline 
\textbf{edges precision} & 0.9985 & 0.9948 & 0.9818 \\ 
\hline 
\textbf{edges recall} & 0.9983 & 0.9946 & 0.9810 \\ 
\hline 
\textbf{sparse codes precision} & 0.7708 & 0.9317 & 0.9244 \\ 
\hline 
\textbf{sparse codes recall} & 0.8001 & 0.9464 & 0.9316 \\ 
\hline 
\end{tabular} 
\label{tab:tests}
\end{table}

Table \ref{tab:tests} shows the mean precision and recall in recovering the graph edges and also the  sparse codes, per graph size, averaged over  100 different graph instances. As expected, the edge recovery accuracy drops slightly with the size of the graph, due to the fact that the number of training signals has not been changed proportionally to the graph size. On the other hand, the recovery performance of sparse codes is much higher in graphs of sizes 50 and 100 than in smaller graphs. We note  that this is probably due to the fact that all tests are performed on training signals constructed from equal number of atoms. For that reason, the overlapping of atoms is much higher in small graphs, making the recovery of sparse codes a more challenging task. 

\begin{figure}[t]
\centering
  \includegraphics[width=1\linewidth]{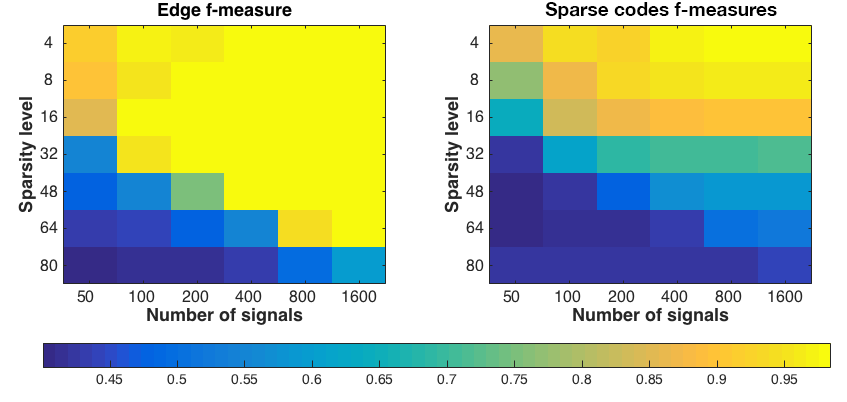}
  \caption{Recovery accuracy depending on sparsity and the number of signals.}
\label{fig:sparsity_nsig}
\end{figure}

Moreover, we have tested the influence of both the sparsity level and the size of the training set on an ER graph with 100 vertices. Fig. \ref{fig:sparsity_nsig} displays the F-measure score  for the recovered edges and the sparse codes. As expected, we can see that a larger number of training signals leads to better recovery. In addition, signals constructed from a smaller number of atoms are more efficient in graph recovery as the atoms are less likely to overlap. Finally, it is worth noting that we have also run experiments in non-ideal settings where the generating kernels of the training signals are not identical to the kernels used in the optimization algorithm. We have observed that our algorithm is pretty robust to such noise which supports its potential towards practical settings.    

%As we assume polynomial coefficients are known, we have also tested robustness of our method to small perturbations in coefficients. Namely, we generated signals from kernels $\hat{g}_1(\lambda) \approx e^{-(2+\epsilon_1) \lambda}$ and $\hat{g}_2(\lambda) \approx 1 - e^{-(1+\epsilon_2) \lambda},$ and witnessed stable results for $\epsilon_1, \epsilon_2 \leq 1$. Stable behaviour also occurred when perturbating polynomial coefficients independently, relatively to their size, $\tilde{\alpha_{sk}}=\alpha_{sk} + \rho_{sk} \alpha_{sk}$, where $\rho_{sk} \in \langle -0.1,0.1 \rangle, \forall s,k.$ In both cases larger number of training signals yielded higher robustness to perturbations.

Next, we compare our model with two state-of-the-art graph learning algorithms, i.e., Dong et al. \cite{Dong_2015} and Segarra et al. \cite{SegarraMMR16}, which have however been designed for smooth graph signal models. In order to have reasonably fair comparisons, we generate piecewise smooth signals, with both generating kernels favouring low frequencies ($\hat{g}_1(\lambda) \approx e^{-2 \lambda}, \hat{g}_2(\lambda) \approx e^{- \lambda}$). We compare these methods on %10
50 ER graphs of size 20, with a probability of an edge of $p=0.3$, and %10 
50 RBF graphs of size 20, and train them on 500 signals. All hyper-parameters are optimized with a grid search. The average F-measure scores are given in Table \ref{tab:compare} and we can see that our method performs better than the other two algorithms. We notice that, when the signals do not necessarily consist of only low-pass kernels, the improvement over these methods is higher as expected.

\begin{table}[h!]
\centering
\caption{Edges F-measures for different methods}
\begin{tabular}{|c|c|c|c|}
\hline 
Type/Method & Our & Dong \cite{Dong_2015}& Segarra \cite{SegarraMMR16} \\ 
\hline 
ER & 1 & 0.9112 & 0.9738\\ 
\hline 
RBF & 1 &  0.9379 & 0.9170 \\ 
\hline     
\end{tabular} 
\label{tab:compare}
\end{table}

% old values
%\begin{table}[h!]
%\centering
%\caption{Edges F-measures for different methods}
%\begin{tabular}{|c|c|c|c|}
%\hline 
%Type/Method & Our & Dong & Segarra \\ 
%\hline 
%ER & 1 & 0.9077 & 0.9709\\ 
%\hline 
%Gauss & 1 &  0.9471 & 0.9020 \\ 
%\hline     
%\end{tabular} 
%\label{tab:compare}
%\end{table}

% New values - 09/09/2016
%Our method:
%ER - 1
%Gauss -1
%
%Dong:
%ER - 0.9112
%Gauss - 0.9379
%
%Segarra
%ER - 0.9738
%Gauss - 0.9170

%\subsection{Real data applications}

%\pascal{Any hope? Or should we stick to synthetic results and promise to have results on real data in the final paper :(}

\section{Conclusions}

In this paper, we have presented a framework for learning graph topologies from signal observations under the assumption that the signals are sparse on a graph-based dictionary. %Specifically, we have proposed an algorithm for learning graphs that enforces sparsity of the graph signals in a generic dictionary that is constructed to be a concatenation of polynomial graph Laplacian matrices that capture different localized patterns on the graph. 
Experimental results confirm the usefulness of the proposed algorithm in recovering a meaningful graph topology and in leading to better data understanding and inference. Even though these results seem promising, %a vital part of most graph-related applications nowadays is a good scalability with the graph size. That is also the main drawback of our approach and 
more efforts are needed to improve the scalability of learning with the graph size, possibly with help of graph reduction and reconstruction methods.

\section{Supplementary materials}
Complementary MATLAB code can be found on:\\ \href{https://github.com/Hermina/GraphLearningSparsityPriors}{https://github.com/Hermina/GraphLearningSparsityPriors}

%.  One interesting course of action might be to explore graph reduction and reconstruction methods and their potential application in approximating inference of large-scale graph structures. 

%We approach the problem of learning graphs from signals with a sparse representation and propose a solution through an EM-like approach. We show that our method successfully recovers the graph topology and give a short analysis of the influence of the number and the structure of training signals. We also show that this model gives meaningful results in realistic graph learning problems. Even though these results seem promising, a vital part of most graph-related applications nowadays is good scalability with graph size. That is also the main drawback of our approach and future efforts must take this problem into account. One interesting course of action might be to explore graph reduction and reconstruction methods and their potential application in approximating inference of large-scale graph structures. (to be updated)

%\vfill\pagebreak

\bibliographystyle{IEEEbib}
\bibliography{mybibfile}

\end{document}